\documentclass[5p]{elsarticle}
\usepackage{subfigure}
\usepackage{booktabs}
\usepackage{amsmath}
\usepackage{subfigure}
\usepackage{makecell}
\newtheorem{myTheo}{Theorem}
\usepackage{url}
\usepackage[colorlinks,linkcolor=red]{hyperref}

\begin{document}
	\begin{frontmatter}
		\title{Naturally Combined Shape-Color Moment Invariants under Affine Transformations\tnoteref{t1,t2}}
		
		\tnotetext[t1]{This work is supported by the National Science Foundation of China under Grant 61379082 and 61227802.}
		\tnotetext[t2]{And this work has been published on Computer Vision and Image Understanding, \href{https://doi.org/10.1016/j.cviu.2017.07.003}{https://doi.org/10.1016/j.cviu.2017.07.003}.}
				
		\author[rvt,rvt2,focal]{Ming Gong\corref{cor2}}
		\ead{yiming10.13@163.com}
		\author[rvt,rvt2]{You Hao\corref{cor1}}
		\ead{haoyou@ict.ac.cn}
		\author[rvt,rvt2]{Hanlin Mo}
		\ead{mohanlin@ict.ac.cn}
		\author[rvt,rvt2]{Hua Li}
		\ead{lihua@ict.ac.cn}
		
		\cortext[cor1]{Co-first author and corresponding author. Tel:+86 10 62600527.}
		\cortext[cor2]{This work is mainly done when Ming Gong pursued her doctor degree in the Institute of Computing Technology. She is now working in Microsoft AI \& Research Group}		
		
		\address[rvt]{Key Laboratory of Intelligent Information Processing, Institute of Computing Technology, Chinese Academy of Sciences, Beijing, China}
		\address[rvt2]{University of Chinese Academy of Sciences, Beijing, China}
		\address[focal]{AI \& Research Group, Microsoft Search Technology Center Asia, Beijing, China}
		\begin{abstract}
			We proposed a kind of naturally combined shape-color affine moment invariants (SCAMI), which consider both shape and color affine transformations simultaneously in one single system. 
			In the real scene, color and shape deformations always exist in images simultaneously. Simple shape invariants or color invariants can not be qualified for this situation. The conventional method is just to make a simple linear combination of the two factors. Meanwhile, the manual  selection of weights is a complex issue. Our construction method is based on the multiple integration framework. The integral kernel is assigned as the continued product of the shape and color invariant cores. It is the first time to directly derive an invariant to dual affine transformations of shape and color. The manual selection of weights is no longer necessary, and both the shape and color transformations are extended to affine transformation group.
			With the various of invariant cores, a set of lower-order invariants are constructed and the completeness and independence are discussed detailedly. A set of SCAMIs, which called SCAMI24, are recommended, and the effectiveness and robustness have been evaluated on both synthetic and real datasets.
		\end{abstract}		
		\begin{keyword}
			Invariants; color; shape; descriptor; dual-affine transformations
		\end{keyword}
	\end{frontmatter}
	
	\section{Introduction}
\begin{figure}
			\centering 
			\includegraphics[width=85mm]{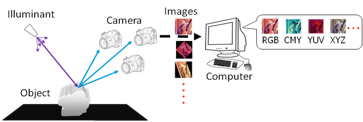}
			\caption{The factors influencing the results during imaging procedure: illumination, camera sensors, and the surface reflective characteristics.} \label{fig:1}
		\end{figure}

		With the development of information technology, the number of images in people's daily life is increasing rapidly. In computer vision, the extraction of image description has been one of the most important tasks. Shape and color are two basic types of information to people's visual cognition and play very important roles in image analysis and understanding. Shape features reflect the position, size, and shape information and color features reflect different spectral reflecting attributes of the object's surface, as shown in Fig.\ref{fig:1}.
		
		The real imaging environment is intricate, which is influenced by different scene illumination, camera sensors, and the reflective characteristics of the objects. Images captured from real scene are always degraded, hence their color and shape are not consistent. This means the color of images are different, and the geometric deformation like scaling, rotation, and skewing are occurring. 
		Various approaches have been proposed to recognize images of the same objects under such geometric and photometric deformations. An effective way is to extract invariant features. 
		
		The 2-D geometric moment invariants were firstly proposed by Hu in 1962 for character recognition \cite{hu1962visual}, which are invariant to similarity transformation. The seven invariants can describe some of features of shape and played important roles in pattern recognition. In \cite{flusser1993pattern}, Suk and Flusser extend the moment invariants from similarity transformation to affine transformation. They are invariant under the affine transformation. The main advantage of invariant features is their invariance under given transformations and there is no need to consider the corresponding deformations in the imaging process. Hence, it was even argued that object recognition is the search for invariants\cite{weiss1993geometric}.

		Color information is useful for object recognition. Many approaches extracting color information are based on color histograms\cite{healey1994using,funt1995color} and color moment\cite{stricker1995similarity}. They make full use of color information but have no color consistency, so are sensitive to illumination changes. Some other color constant descriptors in \cite{li2009illumination, gevers1999color} can deal with the degeneration of illumination. Gong et al.\cite{gong2013moment} proposed a kind of color affine moment invariants which are applicable to more complicated color variance and robust to shape deformation to certain extent. The limitation of the color descriptor is that they do not exploit any spatial information of the object, which leads to vital information lost.
		
\begin{figure}
		\centering 
		\includegraphics[width=85mm]{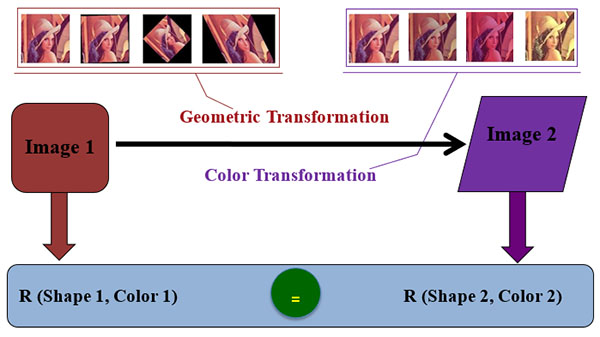}
		\caption{The analysis and construction method of shape and color invariant features.} \label{fig:2}
	\end{figure}

		Lots of efforts have been made on the invariant features which can deal with both shape and color deformations. In common practices, shape and color descriptors are extracted independently, as shown in Fig.\ref{fig:2}. Consequently, they are not very robust in the real condition. It is difficult to combine those two types of information in one descriptor effectively.
		The conventional method is just to make a simple linear combination of the two factors, but manual intervention for the selection of weights is a complex issue. Wang et al. proposed a kind of modified Hu invariant moments\cite{wang2008moment}, which contains both color information and shape information. However, this method is just applicable for the gray-level degradation. 
		Mindru et al. proposed a kind of generalized color moment invariants\cite{mindru2004moment}, extending the moment invariants obtained by Lie group methods detailed in \cite{moons1995foundations,van1995vision}, which considers both shape and color deformations.
		The generalized color moment invariants do not change under geometric affine and photometric diagonal transformations, which get good performance on the experimental datasets.
		For the diagonal transformation of the color with 3-bands (R,G,B), transformation parameters of three color bands can be separated easily, hence the invariants can be constructed in an arbitrary combination of color bands. However, for the affine transformation, it is impossible to separate parameters by different color bands that this method can not be extended to the color affine transformation.
				
		Many works apply invariants into shape's contours to extract shape invariant features\cite{mundy1992goemetric,mundy1994applications}. The limitation of this kind method is that the object contours should be extracted robustly, which is difficult to reach in real scene. Alferez and Wang proposed a method considering both geometric and illumination deformations\cite{alterez1999geometric}. These invariants are based on the sampling curves extracted from the image, which can be the contour of imaged objects or some characteristic curves. This method gets good performance on experimental dataset. The limitation is that the objects' contours should be properly extracted in advance and the weighted average between geometry and illumination invariants also needs manual intervention\cite{alterez1999geometric}.
		
		
		This paper is dedicated to research on Shape-Color Affine Moment Invariants (SCAMI). 
		The construction \linebreak method is based on the multiple integration framework, which can be extended easily along with the various integral kernels. 
		The integral kernel is assigned as the continued product of the shape and color invariant cores, which is based on the geometric invariant primitives.
		
		The main advantage of SCAMI is that they naturally and intrinsically unify shape and color factors together, which is the first time to directly derive an invariant to shape-color dual-affine transformations. Furthermore, the manual selection of weights is no longer necessary. In addition, they can be extended to higher order and dimension easily.
	
		The paper is organized as follows. In Section \ref{sec:background}, the background to construct SCAMI is presented, including the geometric model and the illumination model. In Section \ref{sec:construction}, the invariant constructing framework will be claimed and the derivation of the SCAMI is conducted. Several cases are listed in Section \ref{sec:cases} and the independent of them are tested. In Section \ref{sec:experiments}, the experiment results and conclusion are listed.	
			
	\section{Background}\label{sec:background}
	    The Shape-Color Affine Moment Invariants (SCAMI) are invariant under the shape affine transformation and color affine transformation. 
	    In the real imaging process, the imaging system and conditions are uncertain, which is influenced by different scene illumination, camera sensors, and many other factors.
	    Images captured from real scene are always degraded, hence their color and shape are not consistent. It means that color and shape degradation will bring into the image because of the complex imaging conditions and different systems. In order to deal with these degradations, the transformation mode should be established.
	     	
		\subsection{Shape Transformation} \label{subsec:shape}
		Similar to the imaging mechanism of human's eyes, the camera model is projective transformation from 3D to 2D. When the camera takes images of the same object in \linebreak[4] different viewpoints, the relationship of the images follows the projective transformation. 
		
\begin{figure}
			\centering 
			\includegraphics[width=85mm]{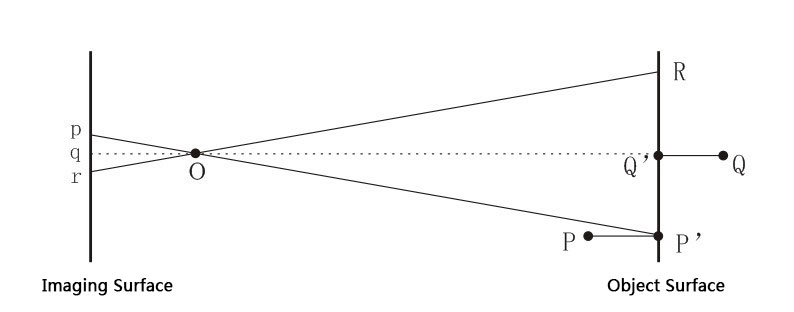}
			\caption{The model of projection approximated by affine transformation.} \label{fig:3}
		\end{figure}

		Fortunately, in the imaging process, when the distance between the camera and the object is far enough, the projective transformation can be well approximated by affine transformation, as shown in the Fig.\ref{fig:3}.
				
		The transformation from $(P,Q)$ to $(p,q)$ is projective transformation. And the similar transformation from \linebreak[4] $(P',Q')$ to $(p,q)$ is affine transformation.
		Therefore, we take affine geometric transformation into consideration. Assume spatial coordinates $(x,y)$ in an image corresponds \linebreak[4] to coordinates $(x',y')$ in another, the expression of the affine transformation can be written as:
			
		\begin{equation}
		\left[ {\begin{array}{*{20}{c}}
		 	{x'}\\
		 	{y'}
		 	\end{array}} \right] = \left( {\begin{array}{*{20}{c}}
		 	{{a_{11}}}&{{a_{12}}}\\
		 	{{a_{13}}}&{{a_{14}}}
		 	\end{array}} \right)\left[ {\begin{array}{*{20}{c}}
		 	x\\
		 	y
		 	\end{array}} \right] + \left[ {\begin{array}{*{20}{c}}
		 	{{t_1}}\\
		 	{{t_2}}
		 	\end{array}} \right] = {M_s} \cdot \left[ {\begin{array}{*{20}{c}}
		 	x\\
		 	y
		 	\end{array}} \right] + {t_0}
		\end{equation}		
		\subsection{Color Transformation}\label{subsec:color}
	    The illumination model models the interaction of light with the surface. But the interaction of light with the surface in the real scenery is too complex to describe. There are illuminating models ranging from simple to complex. 
	    
	    The Kubelka-Munk theory is commonly used as the model to describe the imaging process of the color image. This theory makes a hypothesis that the propagation of light within the material is isotropic, then the properties of the material surface can be described by scattering coefficient and absorption coefficient. This theory can be described as \cite{finlayson2005convex}:	    	    
	    \begin{equation}\label{equ:2}
 	        E(\lambda,\overrightarrow{x}) = e(\lambda,\overrightarrow{x})\cdot(1-\rho_{f}(\overrightarrow{x}))^2\cdot R_{\infty}(\lambda,\overrightarrow{x}) + e(\lambda,\overrightarrow{x})\cdot\rho_{f}(\overrightarrow{x})
	    \end{equation}	    
		where: 
		
		$x$ represents a position of the imaging plane;
		
		$\lambda$ is the wavelength;
		
		$e(\lambda,x)$ is the distribution of light spectrum;
		
		$\rho_f(x)$ is the spectral reflectance in the position x;
		
		$R_{\infty}(\lambda,x)$ is the body reflectance of the object;
		
		$E(\lambda,x)$ is the spectrum distribution in the viewing direction, which can be interpreted as color distribution of the image.
		
		When the optical resolution is high enough, it can be described in a simplified form. Some symbols can be defined as:			
		\begin{align}\label{equ:3}
			&c_{b}(\lambda,\overleftrightarrow{x})\cdot R_{\infty}(\lambda,\overleftrightarrow{x}),  \notag \\
			&c_{i}(\lambda,\overleftrightarrow{x}), \notag \\
			&m_{b}(\overrightarrow{x})=(1-\rho_{f}(\overrightarrow{x}))^2, \notag \\
			&m_{i}(\overrightarrow{x})=\rho_(f)(\overrightarrow{x})
		\end{align}			
		Then the equation can be simplified as:				
		\begin{equation}\label{equ:4}
			E(\lambda,\overrightarrow{x})=m_{b}(\overrightarrow{x})c_{b}(\lambda,\overrightarrow{x})+m_{i}(\overrightarrow{x})c_{i}(\lambda,\overrightarrow{x})
		\end{equation}				
		
\begin{figure}
			\centering 
			\includegraphics[width=85mm]{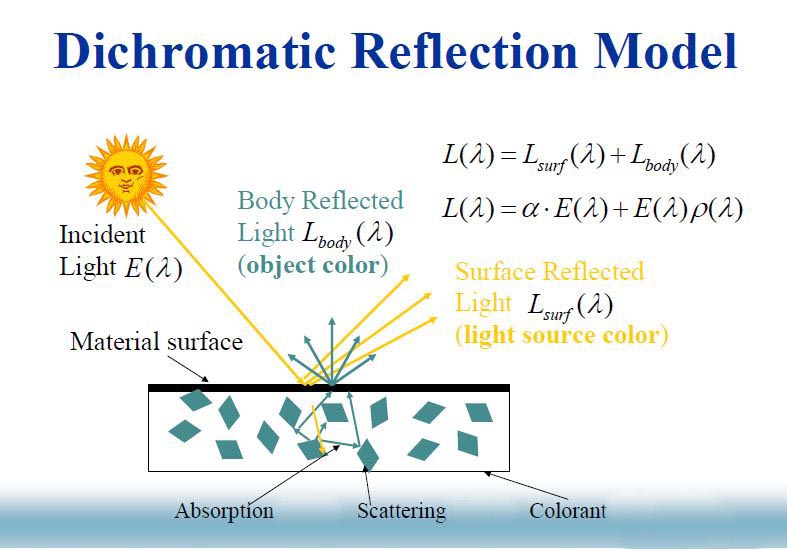}
			\caption{The Dichromatic Reflection Model\cite{wordpress}.} \label{fig:4}
		\end{figure}

		This model is the famous Dichromatic Reflection Model(DRM), which is proposed by Shafer in \cite{shafer1985using}. The model is also commonly used in computer vision, which is presented in Fig.\ref{fig:4}. The reflected light can be split into two parts, one is reflected by the surface and another by the body. In the equation, subscript $i$ represents the surface's part and b the body's part.		
			
		In the \cite{finlayson2005convex}, through adding some constrains to the equation, Finlayson proposed that the transformation between the image $f1$ under the illumination $e1$ and the image $f2$ under the illumination $e2$ can be described with the diagonal-offset model, which can be expressed as:
	    \begin{equation}
		    {f_2} = D \cdot {f_1} + O
	    \end{equation}
		Where $D$ is a 3D diagonal matrix and $det(D)!=0$, which maps the image color captured under illumination $e1$ to that under another $e2$. $O$ represents the offset of each channels. In the $RGB$ color space, this model can be expressed as:
		\begin{equation}\label{equ:6}
		\left( {\begin{array}{*{20}{c}}
			{{R_2}}\\
			{{G_2}}\\
			{{B_2}}
			\end{array}} \right)  = \left( {\begin{array}{*{20}{c}}
			a&0&0\\
			0&b&0\\
			0&0&c
			\end{array}} \right) \cdot \left( {\begin{array}{*{20}{c}}
			{{R_1}}\\
			{{G_1}}\\
			{{B_1}}
			\end{array}} \right) + \left( {\begin{array}{*{20}{c}}
			{{O_R}}\\
			{{O_G}}\\
			{{O_B}}
			\end{array}} \right)
		\end{equation}
		
		Although the diagonal-offset model performs well in describing the indoor scenery, the more complicated models are still needed for describing more complex scenery. In the \cite{geusebroek2001color}, Geusebroek et al. proved that statistically the affine transformation mode is the best linear model to describe the outdoor scenery, which can be described as:
		\begin{equation}\label{equ:7}
		\left( {\begin{array}{*{20}{c}}
			{{R_2}}\\
			{{G_2}}\\
			{{B_2}}
			\end{array}} \right)  = \left( {\begin{array}{*{20}{c}}
			a_{11}&a_{12}&a_{13}\\
			a_{21}&a_{22}&a_{23}\\
			a_{31}&a_{32}&a_{33}
			\end{array}} \right) \cdot \left( {\begin{array}{*{20}{c}}
			{{R_1}}\\
			{{G_1}}\\
			{{B_1}}
			\end{array}} \right) + \left( {\begin{array}{*{20}{c}}
			{{O_R}}\\
			{{O_G}}\\
			{{O_B}}
			\end{array}} \right)
		\end{equation}
		
		Through the equation (\ref{equ:6}) and (\ref{equ:7}), it is obviously presented that the diagonal-offset model is a simple form of the affine transformation model. In this paper, the affine transformation color model is in consideration.		
		
	\section{The Construction Framework of SCAMI}\label{sec:construction}
		The general method to construct SCAMI is introduced in this section. Conveniently, some basic terms used in the method are introduced first. The general idea of the construction framework is to create some invariant geometric primitives as the integral kernel or invariant core. Within the multiple integration framework, we can get the \linebreak[4] SCAMI theoretically. The key point of this method is the selection of geometric invariant core.
		\subsection{Definition of Shape-Color Moment}
		For a piecewise continuous image function $f(x,y)$, the geometric moment of order$(p,q)$ can be defined as:
		\begin{equation}
			M_{pq}^{(f)}=\iint x^py^q\cdot f(x,y)dxdy
		\end{equation}
		where $f(x,y)$ presents the density of gray. For color image $f(x,y)=[R(x,y),G(x,y),B(x,y)]^T$, the shape-color moment, which is called generalized color moment in \cite{mindru2002model}, can be defined as follow:
		\begin{equation} \label{equ:shapecolormoment}
			\begin{split}
			scM_{pq\alpha \beta \gamma } = \iint &{{x^p}{y^q}{{R(x,y)}^\alpha }{{G(x,y) }^\beta }} \cdot {{{B(x,y)}^\gamma }}dxdy
			\end{split}
		\end{equation}
		
		In the shape-color moment form, the geometric center of the image can be expressed as:
		\begin{equation}
			\overline{x} = \frac{{scM_{10000}}}{{scM_{00000}}},\overline{y} = \frac{{scM_{01000}}}{{scM_{00000}}}
		\end{equation}
		
		If the density function of each color channel is piecewise continuous, the moments of all orders exist. $\overline{R},\overline{G},\overline{B}$ are the mean values of each color channel, which can be calculated with equation (\ref{equ:colormean}).
		
		\begin{equation}\label{equ:colormean}
		\overline{c}=\dfrac{\iint c(x,y)dxdy}{\iint dxdy},   c\in \{R,G,B\}
		\end{equation}
		Then the central shape-color moment can be defined as follow:
		\begin{equation}\label{equ:11}
			\begin{split}
				scU_{pq\alpha \beta \gamma } = \iint {{{(x - \overline x )}^p}{{(y - \overline y )}^q}{{(R(x,y) - \overline R )}^\alpha }} \cdot \\
				{{{(G(x,y) - \overline G )}^\beta }{{(B(x,y) - \overline B )}^\gamma }}dxdy
			\end{split}			
		\end{equation}
		
		The advantage of central moment comparing to the general moment is invariant to the translation transformation. Then the invariants constructed by the central moments are translation-invariant. In the following, a set of SCAMIs will be presented in this explicit central shape-color moment form.		
		\subsection{Shape Invariant Primitive}\label{subsec:shapecore}
		In the 2-D geometric space, the central affine transformation matrix can be defined as:		
		\begin{equation}\label{equ:12}
			M_S=\begin{pmatrix}
			a_{11}&a_{12}\\
			a_{21}&a_{22}
			\end{pmatrix}
		\end{equation}		
		
		The area of triangle is relatively invariant under the affine transformation in equation (\ref{equ:12}). The area $A$ of a triangle composed by two point $(x_i,y_i),(x_j,y_j)$ and the origin point $(0,0)$ can be written as:
		\begin{equation}
			A(O,i,j)=\dfrac{1}{2}\cdot (x_iy_j-x_jy_i)
		\end{equation}		
		
		The coefficient $1/2$ have no meaning on the definition of invariants. So for convenience, the coefficient can be ignored.  		
		\begin{equation}
			C_{ij}=(x_iy_j-x_jy_i)=2\cdot A(O,j,j)
		\end{equation}
				
		After affine transformation in equation (\ref{equ:12}), the $C_{ij}$ can be described as:		
		\begin{equation}\label{equ:15}
			C_{ij}'=(x_i'y_j'-x_j'y_i')=J_S\cdot C_{ij}
		\end{equation}					
		Where $J_S$ is the Jacobi determinant of the affine transformation in equation (\ref{equ:12}).		
		\begin{equation}\label{equ:16}
			J_S=\lvert M_S\rvert
		\end{equation}		
					
		So the conclusion comes out obviously that the $C_{ij}$ is a relatively invariant under the affine transformation, which is called shape primitive. Then the shape-invariant core can be constructed with the various primitives composed of different points. The combinations form of the primitives is multiplication as presented in equation (\ref{equ:17}).	
		\begin{equation}\label{equ:17}
			\begin{split}
				saCore&(k,W;{d_1},{d_2},...,{d_k}) \\ 
				&= \underbrace  {A(1,2)A(p,l)...A(r,k)}_{{\rm{there\ are\ }}W{\rm{\ primitives\ }}A}
			\end{split}		    
		\end{equation}		
		Where $k$ is the number of participating points in the shape primitives, $W$ is the number of the primitives and $d_i$ is the multiplicity of the i-th point appearing in the shape core. And $p,l,r,k$ are the points with $p<l, r<k$.  
		
		If we set $saCore'(k,W;d_1,d_2,…,d_k)$ as the corresponding core of the $saCore(k,W;d_1,d_2,…,d_k)$ under the affine transformation in equation (\ref{equ:12}),
		according to the equation (\ref{equ:15}), we can get that
		\begin{equation}
			\begin{split}
				saCore'&(k,W;{d_1},{d_2},...,{d_k}) \\
				&= {\left| {{M_s}} \right|^W} \cdot saCore(k,W;{d_1},{d_2},...,{d_k})
			\end{split}			
		\end{equation}
		
		The shape core is a relatively invariant under affine transformation.
			
		\subsection{Color Invariant Primitive}\label{subsec:colorcore}
		Similarly, in the 3-D geometric space, the central affine transformation matrix can be expressed as:				
		\begin{equation}\label{equ:19}
			M_C=
			\begin{pmatrix}
				a_{11}&a_{12}&a_{13}\\
				a_{21}&a_{22}&a_{23}\\
				a_{31}&a_{32}&a_{33}
			\end{pmatrix}
		\end{equation}		
						
		The volume of a parallelepiped is relatively invariant under the affine transformation in equation (\ref{equ:19}).
		
		As mentioned above, in the 3-D color space of image, the image pixel can be expressed as $(R(x,y),G(x,y),B(x,y))$. For brevity, the pixel $i$ can be written as $(R_i,G_i,B_i)$.
		In the color space, the volume $V$ of a parallelepiped determined by three points $(R_i,G_i,B_i),(R_j,G_j,B_j),(R_k,G_k,B_k)$ and the origin point $(0,0,0)$ can be written as:
				
		\begin{equation}\label{equ:20}
			\begin{array}{l}
				V(i,j,k) = \left| {\begin{array}{*{20}{c}}
				{{R_i}}&{{R_j}}&{{R_k}}\\
				{{G_i}}&{{G_j}}&{{G_k}}\\
				{{B_i}}&{{B_j}}&{{B_k}}\\
				\end{array}} \right|
			\end{array}
		\end{equation}		
		
		Then the $R_i',R_j',R_k'$ can be defined as the corresponding points after the affine transformation in equation (\ref{equ:19}). And the volume after transformation becomes $V'(i,j,k)$ as follow:		
		\begin{equation}\label{equ:21}
			V'(i,j,k)=J_C\cdot V(i,j,k)
		\end{equation}			
		Where $J_C=\left|M_C\right|$ is the Jacobi determinant of the affine transformation in equation (\ref{equ:19}).
		
		It is obviously that $V(i,j,k)$ is relatively invariant under the affine transformation, which is called color primitive. Then the color-affine core can be constructed in the same way as the shape-affine core construction. The color core can be presented as:		
		\begin{equation}\label{equ:22}
			\begin{split}
				ca&Core(n,M;{t_1},{t_2},...,{t_n}) \\
				&= \underbrace {V(1,2,3)V(g,h,i)...V(r,s,n)}_{{\rm{there\ are\ }}M{\rm{\ primitives\ }}V}
			\end{split}			
		\end{equation}	
			
		Similarly, $n$ is the number of points participating in the color primitives, $M$ is the numbers of color primitives and $t_i$ is the multiplicity of the $i$-th pixel appearing in the color core. And $g,h,I,r,s,n$ are the points with $g<h<I,r<s<n$.
		
		If we set $caCore'(n,M;t_1,t_2,…,t_n)$ as the corresponding core of the $caCore(n,M;t_1,t_2,…,t_n)$ under the affine transformation, according to the equation (\ref{equ:21}), we can get that		
		\begin{equation}
			\begin{split}
				caCore'&(n,M;{t_1},{t_2},...,{t_n}) \\
				&= {\left|M_C\right|^M} \cdot caCore(n,M;{t_1},{t_2},...,{t_n})
			\end{split}			
		\end{equation}		
		
		So the color core is also a relatively invariant under affine transformation.
		
		\subsection{Construction}
\begin{figure}
			\centering 
			\includegraphics[width=85mm]{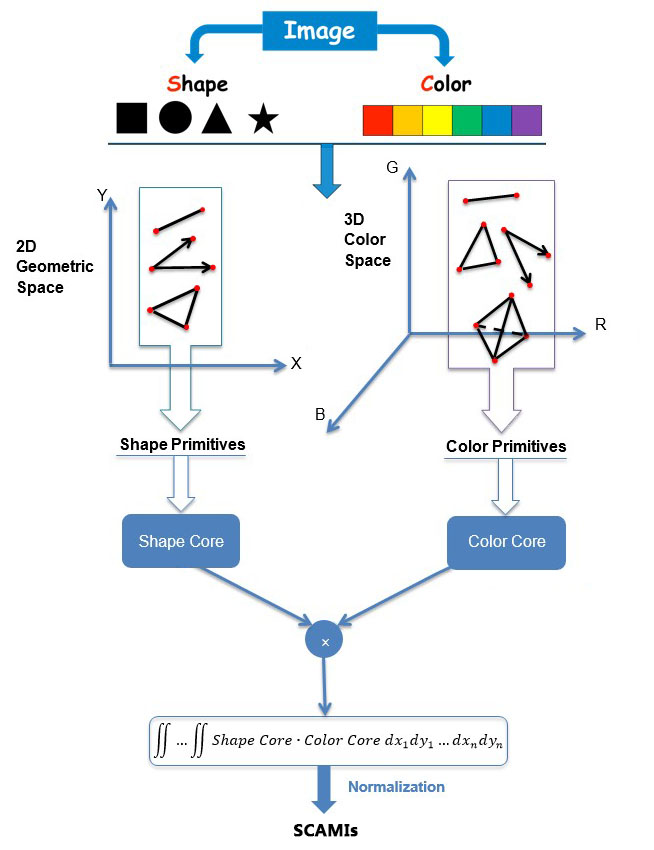}
			\caption{The construction framework based on geometric shape analysis.} \label{fig:5}
		\end{figure}

		Based on the structure of geometric invariant primitives, this part introduces the general method to construct SCAMIs. For the integral kernel $E(n)$, the multiple integral can be defined as:		
		\begin{equation}
		    I(E(n)) = \iint {...}\iint {E(n)d{x_1}d{y_1}...d{x_n}d{y_n}}
	    \end{equation}				
		Where $E(n)$ is an expression of $n$ points and $(x_i,y_i)$ are the coordinates of the $i$-th points. And this is a $n$-ple integral for the expression $E(n)$. Based on the multiple integral and the geometric invariant primitives, the integral-based method to construct Shape-Color Affine Moment Invariants is proposed. In the Fig.\ref{fig:5}, the method is presented visually.		
		Based on it, the theorem is shown as:
		
		\begin{myTheo}
		 	The shape-color moment invariants are constructed as equation (\ref{equ:scami}).
		\end{myTheo}

		\begin{equation} \label{equ:scami}
		\frac{
			\left\{\begin{aligned}
			I(&saCore({N_1},{m_1};{d_1},{d_2},...,{d_{{N_1}}}) \cdot \\
			 &caCore({N_2},{m_2};{t_1},{t_2},...,{t_{{N_2}}}))  \\
			\end{aligned}\right\}
		}
		{
			\left\{\begin{aligned}
			&I(saCore(1,0))^{\max ({N_1},{N_2}) + {m_1} - 3{m_2}/2} \cdot  \\ 
			&I(caCore(3,2;2,2,2))^{{m_2}/2}
			\end{aligned}\right\}
		}
		\end{equation}
		\textbf{Proof.} According to the properties of shape and color core under affine transformation in Section \ref{subsec:shapecore} and \ref{subsec:colorcore}.		
		\begin{align}\label{equ:26}
			&\mathrel{\phantom{=}}I(saCore'({N_1},{m_1};{d_1},{d_2},...,{d_{{N_1}}}) \cdot \notag \\ 
			&\mathrel{\phantom{=}}caCore'({N_2},{m_2};{t_1},{t_2},...,{t_{{N_2}}})) \notag \\
			&=\iint ... \iint saCore'({N_1},{m_1};{d_1},{d_2},...,{d_{{N_1}}}) \cdot \notag \\
			&\mathrel{\phantom{=\iint ... \iint }}caCore'({N_2},{m_2};{t_1},{t_2},...,{t_{{N_2}}}) \notag \\ 
			&\mathrel{\phantom{=\iint ... \iint }}dx_1'dy_1'...dx_N'dy_N' \notag \\
			&={\left| {{M_s}} \right|^{{m_1} + N}} \cdot {\left| {{M_c}} \right|^{{m_2}}} \cdot \notag \\
			&\mathrel{\phantom{=}}\iint ... \iint saCore({N_1},{m_1};{d_1},{d_2},...,{d_{{N_1}}}) \cdot \notag \\ &\mathrel{\phantom{=\iint ... \iint }}caCore({N_2},{m_2};{t_1},{t_2},...,{t_{{N_2}}}) \notag \\
			&\mathrel{\phantom{=\iint ... \iint }}d{x_1}d{y_1}...d{x_N}d{y_N} \notag \\
			&={\left| {{M_s}} \right|^{{m_1} + \max ({N_1},{N_2})}} \cdot {\left| {{M_c}} \right|^{{m_2}}} \cdot \notag \\
			&\mathrel{\phantom{=}}I(saCore({N_1},{m_1};{d_1},{d_2},...,{d_{{N_1}}}) \cdot \notag \\
			&\mathrel{\phantom{=}}caCore({N_2},{m_2};{t_1},{t_2},...,{t_{{N_2}}}))
		\end{align}

		Under the transformations of $M_S$ and $M_C$, the denominator of equation (\ref{equ:scami}) will be		
		\begin{align}\label{equ:27}
			&\mathrel{\phantom{=}}I{(saCore'(1,0))^{\max ({N_1},{N_2}) + {m_1} - 3{m_2}/2}} \notag \\
			&= {(\iint dx'dy')^{\max ({N_1},{N_2}) + {m_1} - 3{m_2}/2}} \notag \\
			&= {(\left| {{M_s}} \right|\iint dxdy)^{\max ({N_1},{N_2}) + {m_1} - 3{m_2}/2}} \notag \\
			&= {\left| {{M_s}} \right|^{\max ({N_1},{N_2}) + {m_1} - 3{m_2}/2}} \cdot \notag \\ &\mathrel{\phantom{=}}I{(saCore(1,0))^{\max ({N_1},{N_2}) + {m_1} - 3{m_2}/2}}
		\end{align}		
		\begin{align}\label{equ:28}
			&\mathrel{\phantom{=}}I{(caCore'(3,2;2,2,2))^{{m_2}/2}} \notag \\
			&= (\iint ... \iint caCore'(3,2;2,2,2) \cdot \notag \\ 
			&\mathrel{\phantom{= (\iint ... \iint }}dx_1'dy_1'dx_2'dy_2'dx_3'dy_3')^{{m_2}/2} \notag \\
			&= ({\left| {{M_c}} \right|^2} \cdot {\left| {{M_s}} \right|^3} \cdot \iint ... \iint caCore(3,2;2,2,2) \cdot \notag \\
			&\mathrel{\phantom{= ({\left| {{M_c}} \right|^2} \cdot {\left| {{M_s}} \right|^3} \cdot \iint ... \iint}}d{x_1}d{y_1}d{x_2}d{y_2}d{x_3}d{y_3})^{{m_2}/2} \notag \\
			&= \left| {{M_c}} \right|^{{m_2}} \cdot \left| {{M_s}} \right|^{3{m_2}/2} \cdot I{(caCore(3,2;2,2,2))^{{m_2}/2}}
		\end{align}	
					
		Therefore, it is proved that equation (\ref{equ:scami}) is shape-color moment invariants to dual affine transformations.	This ends the proof.
		
		Theorem 1 gives the general form of the SCAMI. It is the first time to directly derive an invariant to dual affine transformations of shape and color, which is the most complicated linear transformation. In this framework, the invariants can be extended to higher order and higher dimensions, and generate infinite numbers of SCAMI theoretically. And the infinite of invariant cores makes the method to be extensible conveniently. In order to instantiate the invariant, various invariant cores will be applied into equation (\ref{equ:scami}), which will be demonstrated in Section \ref{sec:cases}.
		
	\section{SCAMI24 in Lower Order}\label{sec:cases}
		\subsection{The Shape-Color Invariant Cores}
		According to the definition of moment invariant, the high order moments involved in the invariant functions are susceptible to noise in the images. 
		This has also been illuminated in many works\cite{mindru2004moment,mindru2002model}. Another limitation of the invariants with high order is the computation complexity. Under these considerations, we choose all the shape cores whose orders are no more than 4 and participation points are also no more than 4. The orders of color cores are no more than 2 and the participation points are limited as 3.
		
\begin{figure}
			\centering 
			\subfigure[]{
				\scalebox{0.6}{\includegraphics{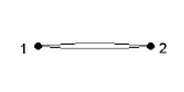}}
				\label{fig:graphmethoda}
			}
			\subfigure[]{
				\scalebox{0.6}{\includegraphics{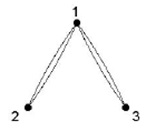}}
				\label{fig:graphmethodb}			
			}
			\caption{The graphs corresponding to examples of shape cores.} 
			\label{fig:graphmethod}
		\end{figure}

\begin{figure}
			\centering 
			\subfigure[]{\includegraphics[width=13mm]{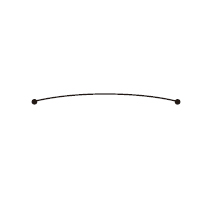}
				\label{fig:sc211}}
			\subfigure[]{\includegraphics[width=13mm]{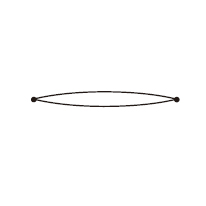} 
				\label{fig:sc212}}
			\subfigure[]{\includegraphics[width=13mm]{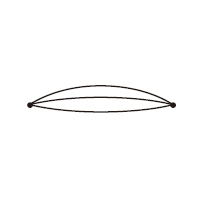} 
				\label{fig:sc213}}
			\subfigure[]{\includegraphics[width=13mm]{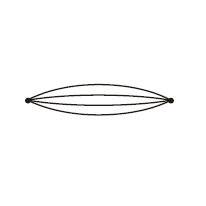} 
				\label{fig:sc214}}
			\subfigure[]{\includegraphics[width=13mm]{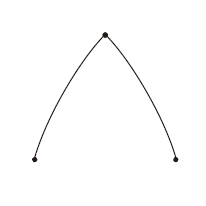} 
				\label{fig:sc321}}
			\subfigure[]{\includegraphics[width=13mm]{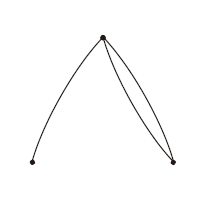} 
				\label{fig:sc322}}
			\subfigure[]{\includegraphics[width=13mm]{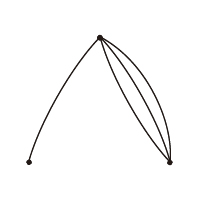} 
				\label{fig:sc323}}
			\subfigure[]{\includegraphics[width=13mm]{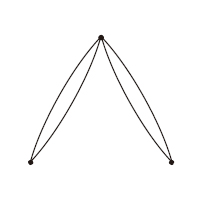} 
				\label{fig:sc324}}
			\subfigure[]{\includegraphics[width=13mm]{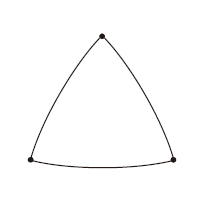} 
				\label{fig:sc331}}
			\subfigure[]{\includegraphics[width=13mm]{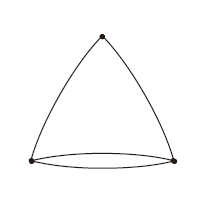} 
				\label{fig:sc332}}
			\subfigure[]{\includegraphics[width=13mm]{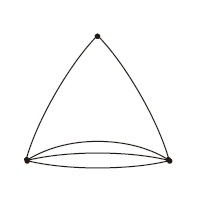} 
				\label{fig:sc333}}
			\subfigure[]{\includegraphics[width=13mm]{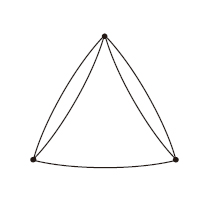} 
				\label{fig:sc334}}
			\subfigure[]{\includegraphics[width=13mm]{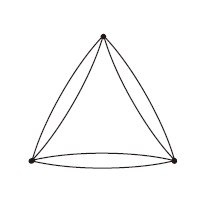} 
				\label{fig:sc335}}
			\subfigure[]{\includegraphics[width=13mm]{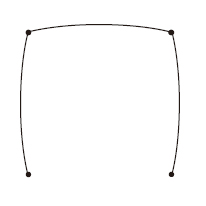}
				 \label{fig:sc4311}}
			\subfigure[]{\includegraphics[width=13mm]{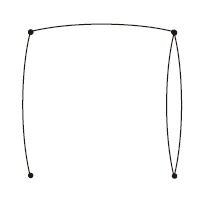}
				 \label{fig:sc4312}}
			\subfigure[]{\includegraphics[width=13mm]{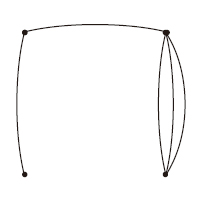}
				 \label{fig:sc4313}}
			\subfigure[]{\includegraphics[width=13mm]{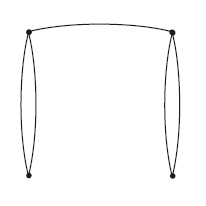}
				 \label{fig:sc4314}}
			\subfigure[]{\includegraphics[width=13mm]{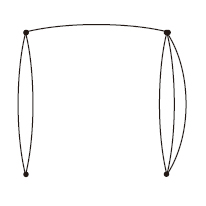}
				 \label{fig:sc4315}}
			\subfigure[]{\includegraphics[width=13mm]{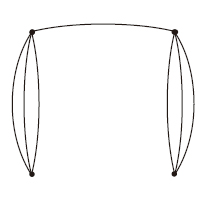}
				 \label{fig:sc4316}}
			\subfigure[]{\includegraphics[width=13mm]{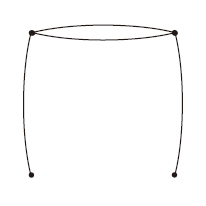}
				 \label{fig:sc4317}}
			\subfigure[]{\includegraphics[width=13mm]{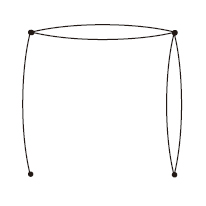}
				 \label{fig:sc4318}}
			\subfigure[]{\includegraphics[width=13mm]{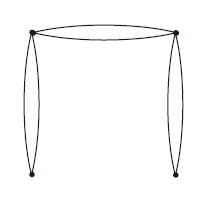}
				 \label{fig:sc4319}}
			\subfigure[]{\includegraphics[width=13mm]{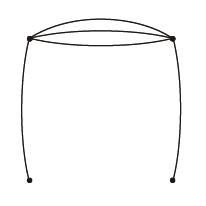}
				 \label{fig:sc43110}}
			\subfigure[]{\includegraphics[width=13mm]{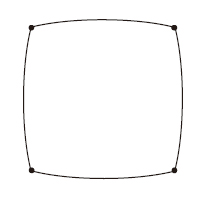}
				 \label{fig:sc4411}}
			\subfigure[]{\includegraphics[width=13mm]{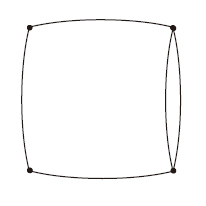}
				 \label{fig:sc4412}}
			\subfigure[]{\includegraphics[width=13mm]{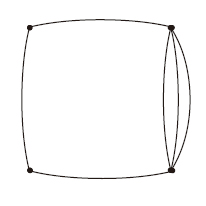}
				 \label{fig:sc4413}}
			\subfigure[]{\includegraphics[width=13mm]{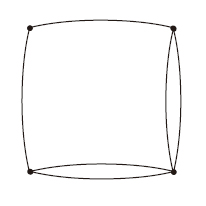}
				 \label{fig:sc4414}}
			\subfigure[]{\includegraphics[width=13mm]{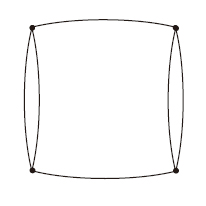}
				 \label{fig:sc4415}}
			\subfigure[]{\includegraphics[width=13mm]{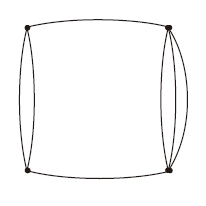}
				 \label{fig:sc4416}}
			\subfigure[]{\includegraphics[width=13mm]{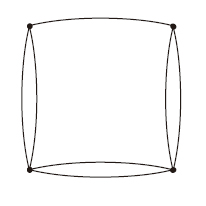}
				 \label{fig:sc4417}}
			\subfigure[]{\includegraphics[width=13mm]{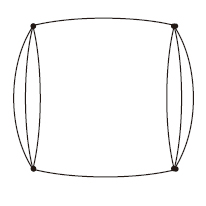}
				 \label{fig:sc4418}}
			\subfigure[]{\includegraphics[width=13mm]{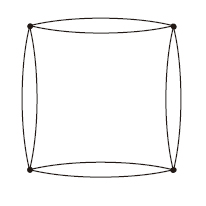}
				 \label{fig:sc4419}}
			\subfigure[]{\includegraphics[width=13mm]{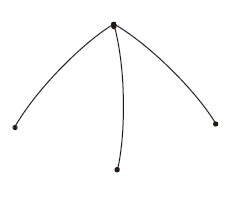}
				 \label{fig:sc4321}}
			\subfigure[]{\includegraphics[width=13mm]{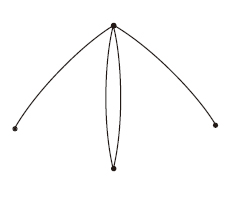}
				 \label{fig:sc4322}}
			\subfigure[]{\includegraphics[width=13mm]{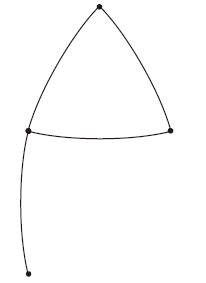}
				 \label{fig:sc4421}}
			\subfigure[]{\includegraphics[width=13mm]{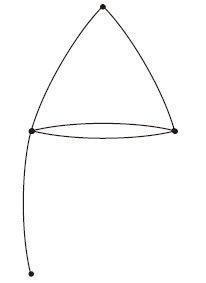}
				 \label{fig:sc4422}}
			\subfigure[]{\includegraphics[width=13mm]{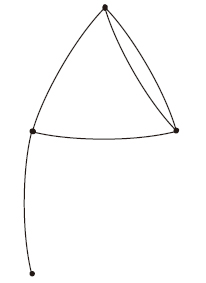}
				 \label{fig:sc4423}}
			\subfigure[]{\includegraphics[width=13mm]{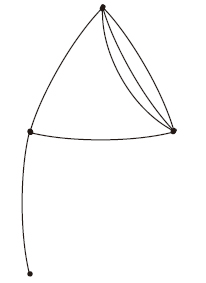}
				 \label{fig:sc4424}}
			\subfigure[]{\includegraphics[width=13mm]{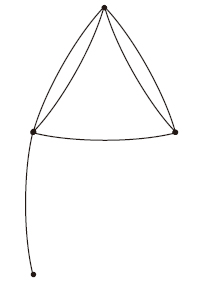}
				 \label{fig:sc4425}}
			\subfigure[]{\includegraphics[width=13mm]{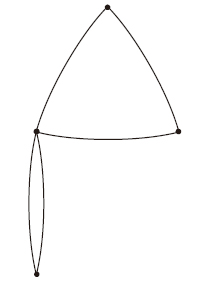}
				 \label{fig:sc4426}}
			\subfigure[]{\includegraphics[width=13mm]{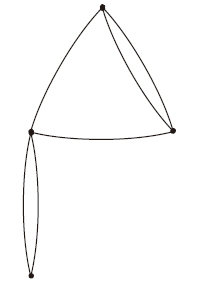}
				 \label{fig:sc4427}}
			\subfigure[]{\includegraphics[width=13mm]{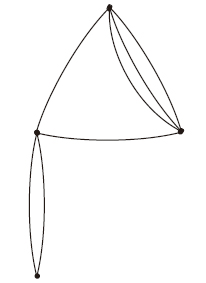}
				 \label{fig:sc4428}}
			\subfigure[]{\includegraphics[width=13mm]{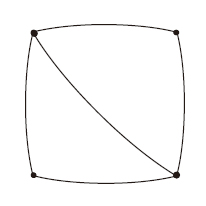} 
				\label{fig:sc451}}
			\subfigure[]{\includegraphics[width=13mm]{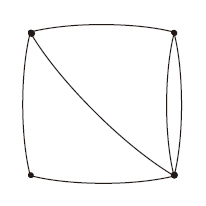} 
				\label{fig:sc452}}
			\subfigure[]{\includegraphics[width=13mm]{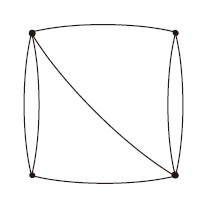} 
				\label{fig:sc453}}
			\subfigure[]{\includegraphics[width=13mm]{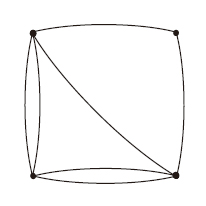} 
				\label{fig:sc454}}
			\subfigure[]{\includegraphics[width=13mm]{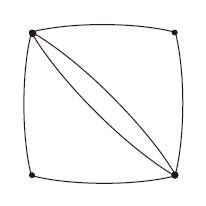} 
				\label{fig:sc455}}
			\subfigure[]{\includegraphics[width=13mm]{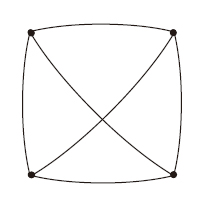} 
				\label{fig:sc461}}
			\subfigure[]{\includegraphics[width=13mm]{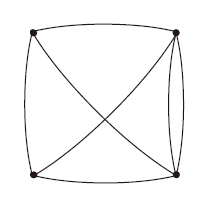} 
				\label{fig:sc462}}
			\subfigure[]{\includegraphics[width=13mm]{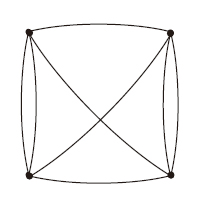} 
				\label{fig:sc463}}
			\caption{The 50 low order shape cores.} 
			\label{fig:shapecores}
		\end{figure}

		Suk et al. \cite{suk2004graph} proposed a graph method to generate shape cores of moment invariants which can represent the shape cores clearly. The examples of this method is shown in Fig.\ref{fig:graphmethod}.
		The point 1,2,3 represent to the participation points in the shape cores, and the edge between two points represent to the primitive constructed by the two end points. Thus, Fig \ref{fig:graphmethoda} represents the shape core $(x_1y_2-x_2y_1)^2$, Fig \ref{fig:graphmethodb} represents the shape core $(x_1y_2-x_2y_1)^2 * (x_1y_3-x_3y_1)^2$. Then all the shape cores whose orders are no more than 4 and participation points are also no more than 4 can be represented as graphs as shown in Fig \ref{fig:shapecores}. There are 50 shape cores in total.

		There are 2 color cores in total whose order no more than 2 and participation points no more than 3, shown in equation (\ref{equ:cacore1}) and (\ref{equ:cacore2}).
		\begin{align} \label{equ:cacore1}
			caCore(3,2;2,2,2) &=  V(1,2,3)^2 \notag \\
			&=({R_3}{G_1}{B_2} - {R_3}{G_2}{B_1} + \notag \\
			&\mathrel{\phantom{=}}{R_2}{G_3}{B_1} - {R_1}{G_3}{B_2} + \notag \\
			&\mathrel{\phantom{=}}{R_1}{G_2}{B_3} - {R_2}{G_1}{B_3})^2
		\end{align}
		\begin{align} \label{equ:cacore2}
			caCore(3,1;1,1,1) &=  V(1,2,3) \notag \\
			&=({R_3}{G_1}{B_2} - {R_3}{G_2}{B_1} + \notag \\
			&\mathrel{\phantom{=}}{R_2}{G_3}{B_1} - {R_1}{G_3}{B_2} + \notag \\
			&\mathrel{\phantom{=}}{R_1}{G_2}{B_3} - {R_2}{G_1}{B_3})
		\end{align}
		
		Finally, 50 shaped cores are combined with the 2 color cores, forming a total 100 shape-color invariant cores.
		The 100 shape-color cores are taken into the equation (\ref{equ:scami}), respectively, to obtain 100 shape-color dual-affine invariants.
		  
		\subsection{Independence of the SCAMIs}
		
		With the selection of the shape-color cores, we find all the low order moments with certain limitation.
		It is easily to expand them into the rational expressions of the shape-color moments in equation  (\ref{equ:shapecolormoment}). Unfortunately, some SCAMIs are useless because they are equal to zero.
		Another important factor for the SCAMIs' selection is the dependencies among the invariants. 
		It is important for applications to verify the independence between invariants. And it is a complicated task because the invariants are functions in the form of high degree polynomials in moments.		
		There are several kinds of criterion for the selection of SCAMIs. We can group them into three classes.
		
		\textit{Zero invariants.}
		In the first case, the expansion of some SCAMIs are equal to zero. These invariants can not be used as the image descriptors, because they are all invariant for any images the are applied. There are 30 SCAMIs of this kind in total.
		
		For another case, as demonstrated in \cite{suk2004graph}, when using central moments, all first-order moments are zero by definition and, consequently, such invariants are zero, too. For the shape-color moments in equation (\ref{equ:shapecolormoment}), this theory is also true. The first-order shape-color moments are also equal to zero when using central moments, no matter first-order shape moments or first-order color moments. In this case, there are 32 SCAMIs are equal to zero in total. 

		\textit{Linear combinations.}
		The dependencies among invariants is common, some invariants may be equal to another, or equal to the linear combination of some other invariants. After excluding the invariants equaling to zero, there are 38 SCAMIs left. All possible linear combinations of the SCAMIs are checked. The number of linear combinations is so much that it is difficult to enumerate. Fortunately, a necessary condition for linear dependence is that all invariants should have the same numbers of shape-color moments of the same order. This will greatly reduce the possible situation. Finally, four linear dependencies are found in total. Hence there are 34 SCAMIs which are linear independent remained.
		
		\textit{Functional dependencies.}	
		"Linear Independent" does not mean "Absolutely Independent". There may exists functional relationship with higher-order polynomial dependencies among these SCAMIs.
		Verifying the independence between them is important for applications. But that is a complicated task, for the invariants are functions in the form of high degree polynomials in moments. 
		
		Fortunately, there is a theorem as follow:
		
		\begin{myTheo}
			The condition for dependence of n functions of n+p variables is that every determinant of order n formed from the matrix of the first partial derivatives vanish identically.\cite{brown1935functional}
		\end{myTheo}
		
		It means that N functions are independent as long as there exists a nonsingular n-order determinant that is not zero. 
		
		\begin{equation}\label{equ:38}
		J = \left| {\begin{array}{*{20}{c}}
			{\frac{{\partial {f_1}}}{{\partial {x_1}}}}&{\frac{{\partial {f_1}}}{{\partial {x_2}}}}& \cdots &{\frac{{\partial {f_1}}}{{\partial {x_{n + p}}}}}\\
			{\frac{{\partial {f_2}}}{{\partial {x_1}}}}&{\frac{{\partial {f_2}}}{{\partial {x_2}}}}& \cdots &{\frac{{\partial {f_2}}}{{\partial {x_{n + p}}}}}\\
			\vdots & \vdots & \ddots & \vdots \\
			{\frac{{\partial {f_n}}}{{\partial {x_1}}}}&{\frac{{\partial {f_n}}}{{\partial {x_2}}}}& \cdots &{\frac{{\partial {f_n}}}{{\partial {x_{n + p}}}}}
			\end{array}} \right|
		\end{equation}
		
		In the definition of 34 linear independent SCAMIs, there are 150 shape-color moments, like $scU_{00000}$, $scU_{00001}$, and $scU_{22110}$, in total. So 34 functions with 150 variables are established. And a $34\times150$ matrix of the first partial derivatives needs to be checked. The complexity to calculate an $n$-order determinant is $n!\cdot(n-1)$, and the number of $n$-order sub-matrix is $C_{n+p}^n$. Thus the complexity of this problem is $O(n!\cdot n)$. For our existing experimental environment, the problem is complex enough to get the result within an acceptable time. 
		
		\begin{figure}
			\centering 
			\subfigure[]{\includegraphics[width=85mm]{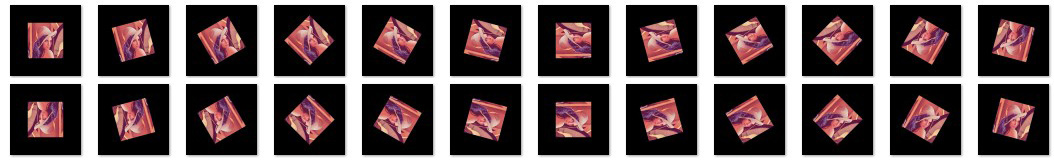}
				\label{fig:exp1rotate}}
			\subfigure[]{\includegraphics[width=85mm]{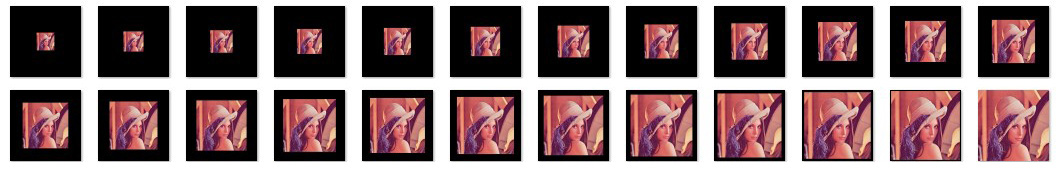}
				\label{fig:exp1scale}}
			\subfigure[]{\includegraphics[width=85mm]{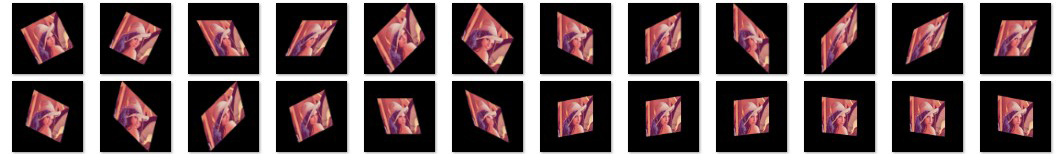}
				\label{fig:exp1affine}}
			\subfigure[]{\includegraphics[width=85mm]{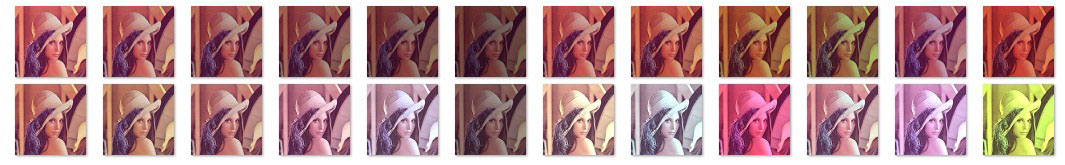}
				\label{fig:exp1color}}			
			\caption{The sample images after artificial deformations: (a) rotation; (b) scale; (c) shape affine; (d) color affine.} 
			\label{fig:exp1}
		\end{figure}
		
		Fortunately, not every SCAMI contains all the 150 variables and most of them contain 20 to 60. Only three pairs SCAMIs contain exactly the same variables. Therefore, the matrix is very sparse with most of entries are zero.
		The computational complexity will be greatly reduced. The rank of the matrix is calculated in Maple2015 and the result is 34. Then we can get the conclusion that the 34 invariant above are independent.
		
		\subsection{Invariant Capability  Analysis}
		In order to verify the performance of each invariant, we designed experiments under different transformations \linebreak synthetically. The experiments aim at evaluating the robustness and discriminant power of the SCAMIs under deformation from the ideal conditions. Artificial shape and color transformations were applied to sample images, including rotation, scale, affine, and color affine transformations, which are shown in Fig. (\ref{fig:exp1}).			
		
		In addition, we generate 400 deformed images with multiple transformations as a final evaluation.
		The mean relative error (MRE) are calculated for each SCAMI on all datasets.
		The results of these five group experiments are listed in Table \ref{table:exp1}.
		\begin{table}
			\centering
			\small
			\caption{The MRE of each SCAMI for different deformations}
			\label{table:exp1}
			\begin{tabular}{cccccc}
				\toprule
				\textit{SCAMI}& Rotate & Scale & Affine & Color  & Multiple \\
				\midrule
				(1)  & 0.0034  & 0.0232 & 0.0012  & 0.1546 & 0.0228   \\
				(2)  & 0.0059  & 0.0328 & 0.0033  & 0.1999 & 0.0430   \\
				(3)  & 0.0096  & 0.0286 & 0.0007  & 0.1467 & 0.0571   \\
				(4)  & 0.0077  & 0.0280 & 0.0007  & 0.1538 & 0.0632   \\
				(5)  & 0.0058  & 0.0234 & 0.0016  & 0.2196 & 0.0350   \\
				(6)  & 0.0150  & 0.0345 & 0.0640  & 0.2815 & \textbf{0.1130}   \\
				(7)  & 0.0158  & 0.7251 & 0.0640  & 0.2815 & 0.1042   \\
				(8)  & 0.0004  & 0.0236 & 0.0300  & 0.1480 & 0.0254   \\
				(9)  & 0.0463  & 0.1254 & 0.3046  & 0.2053 & \textbf{0.4567}   \\
				(10) & 0.7730  & 0.1039 & 0.3789  & 0.2455 & \textbf{0.4813}   \\
				(11) & 0.0256  & 0.0231 & 0.1005  & 0.1443 & 0.0293   \\
				(12) & 0.0242  & 0.0231 & 0.1029  & 0.1481 & 0.0266   \\
				(13) & 0.0084  & 0.0284 & 0.0311  & 0.1499 & 0.0612   \\
				(14) & 0.0069  & 0.0274 & 0.0321  & 0.1584 & 0.0685   \\
				(15) & 0.0019  & 0.0236 & 0.0879  & 0.3518 & 0.0450   \\
				(16) & 0.0223  & 0.0043 & 0.0003  & 0.0133 & 0.0232   \\
				(17) & 0.0373  & 0.0063 & 0.0006  & 0.0203 & 0.0324   \\
				(18) & 0.0583  & 0.0457 & 0.0019  & 0.1833 & 0.1022   \\
				(19) & 0.0155  & 0.0472 & 0.0035  & 0.1839 & 0.0920   \\
				(20) & 0.0163  & 0.6441 & 0.0035  & 0.1839 & 0.0796   \\
				(21) & 0.0385  & 0.7111 & 0.0006  & 0.0190 & 0.0594   \\
				(22) & 0.1749  & 0.0746 & 0.0182  & 0.6362 & \textbf{0.6068}   \\
				(23) & 0.0519  & 0.0122 & 0.0008  & 0.0205 & 0.0482   \\
				(24) & 0.0626  & 0.0458 & 0.1450  & 0.2363 & \textbf{0.1205}   \\
				(25) & 0.0698  & 0.0251 & 0.5917  & 0.1748 & \textbf{0.3154}   \\
				(26) & 0.0050  & 0.0502 & 0.3360  & 0.2327 & \textbf{0.1186}   \\
				(27) & 0.0456  & 0.0090 & 0.0326  & 0.0203 & 0.0410   \\
				(28) & 0.0327  & 0.0064 & 0.0346  & 0.0159 & 0.0335   \\
				(29) & 0.0260  & 0.0075 & 0.1046  & 0.0179 & 0.0458   \\
				(30) & 0.0549  & 0.0425 & 0.4241  & 0.3137 & \textbf{0.1238}   \\
				(31) & 0.0788  & 0.0258 & 0.6537  & 0.2325 & \textbf{0.3605}   \\
				(32) & 0.0372  & 0.0096 & 0.1037  & 0.0231 & 0.0502   \\
				(33) & 0.1558  & 0.0925 & 0.5639  & 0.7035 & \textbf{0.5084}   \\
				(34) & 0.0470  & 0.0109 & 0.0299  & 0.0196 & 0.0471   \\
				\bottomrule 
			\end{tabular}
		\end{table}

		As shown in the results, the characteristics of each SCAMI for different deformations are listed clearly. For the shape deformations listed in the first three columns of the table, the MRE of most SCAMIs is less than 10\%. Several SCAMIs are not accurate enough for particular transformation. Such as SCAMI(7), SCAMI(20), and SCAMI(12) get high MRE for the scaling transformation but low MRE for others. This is determined by the characteristics of invariant cores that make up the SCAMIs. For the color deformations listed in the forth columns of the table, the SCAMIs' MRE are not as accurate as that under shape deformations. This is mainly due to the discrete calculation for color affine transformations. The image values are distributed between 0 and 255. The color affine transformation may map some of the different values into the same, even outside of the range. This will result in certain loss of information, especially under severe transformation as shown in Fig. \ref{fig:exp1color}. But most of them are less than 20\%, which does not affect their effectiveness. The advantage of them compared to other color descriptors will be demonstrated in the experiments.
		
		For a comprehensive evaluation, the results with multiple deformations listed in the fifth column are considered as the criteria. Finally, 24 effective SCAMIs, whose MRE is no more than 10\% in the experiment with multiple deformations, are selected, which can be used as the basic descriptor in the process of analyzing and understanding images. 
		
		For the sake of clarity, The serial numbers are rearranged from 1 to 24. The invariant cores chosen for the 24 SCAMIs are presented in Table \ref{table:1}, where $V(1, 2, 3)$ is the volume of the parallelepiped consisting of 3 points, which is clarified in equation (\ref{equ:20}).
		
		\begin{table}
						\renewcommand\arraystretch{1.18}
			\centering
			\caption{The Color and Shape Cores for the $SCAMI24$}
			\label{table:1}
			\begin{tabular}{cll}
				\toprule
				SCAMI & Color Core & Shape Core \\
				\midrule
				(1) & $V{(1,2,3)^2}$ & ${({x_1}{y_2} - {x_2}{y_1})^2}$ \\
				(2) & $V{(1,2,3)^2}$ & ${({x_1}{y_2} - {x_2}{y_1})^4}$ \\
				(3) & $V{(1,2,3)^2}$ & $({x_1}{y_2} - {x_2}{y_1})({x_1}{y_3} - {x_3}{y_1})$ \\
				(4) & $V{(1,2,3)^2}$ & $({x_1}{y_2} - {x_2}{y_1}){({x_1}{y_3} - {x_3}{y_1})^3}$ \\
				(5) & $V{(1,2,3)^2}$ & ${({x_1}{y_2} - {x_2}{y_1})^2}{({x_1}{y_3} - {x_3}{y_1})^2}$ \\
				(6) & $V{(1,2,3)^2}$ & ${({x_1}{y_2} - {x_2}{y_1})^2}{({x_1}{y_3} - {x_3}{y_1})^2}\cdot $\\
				&&${({x_2}{y_3} - {x_3}{y_2})^2}$ \\
				(7) & $V{(1,2,3)^2}$ & ${({x_1}{y_2} - {x_2}{y_1})^2}{({x_2}{y_3} - {x_3}{y_2})^2}
				\cdot $\\
				&&${({x_3}{y_4} - {x_4}{y_3})^2}$ \\
				(8) & $V{(1,2,3)^2}$ & $({x_1}{y_2} - {x_2}{y_1})({x_2}{y_3} - {x_3}{y_2})\cdot $\\
				&&$({x_3}{y_4} - {x_4}{y_3})({x_1}{y_4} - {x_4}{y_1})$ \\
				(9) & $V{(1,2,3)^2}$ & $({x_1}{y_2} - {x_2}{y_1})({x_2}{y_3} - {x_3}{y_2})\cdot $\\
				&&$({x_3}{y_4} - {x_4}{y_3}){({x_1}{y_4} - {x_4}{y_1})^3}$ \\
				(10) & $V{(1,2,3)^2}$ & $({x_1}{y_2} - {x_2}{y_1}){({x_2}{y_3} - {x_3}{y_2})^3}\cdot $\\
				&&$({x_3}{y_4} - {x_4}{y_3}){({x_1}{y_4} - {x_4}{y_1})^3}$ \\
				(11) & $V{(1,2,3)^2}$ & ${({x_1}{y_2} - {x_2}{y_1})^2}{({x_2}{y_3} - {x_3}{y_2})^2}\cdot $\\
				&&${({x_3}{y_4} - {x_4}{y_3})^2}{({x_1}{y_4} - {x_4}{y_1})^2}$ \\
				(12) & $V(1,2,3)$ & $({x_1}{y_2} - {x_2}{y_1}){({x_1}{y_3} - {x_3}{y_1})^2}$ \\
				(13) & $V(1,2,3)$ & $({x_1}{y_2} - {x_2}{y_1})({x_1}{y_3} - {x_3}{y_1})\cdot $\\
				&&$({x_2}{y_3} - {x_3}{y_2})$ \\
				(14) & $V(1,2,3)$ & $({x_1}{y_2} - {x_2}{y_1})({x_1}{y_3} - {x_3}{y_1})\cdot $\\
				&&${({x_2}{y_3} - {x_3}{y_2})^3}$ \\
				(15) & $V(1,2,3)$ & ${({x_1}{y_2} - {x_2}{y_1})^2}{({x_1}{y_3} - {x_3}{y_1})^2}\cdot $\\
				&&$({x_2}{y_3} - {x_3}{y_2})$ \\
				(16) & $V(1,2,3)$ & ${({x_1}{y_2} - {x_2}{y_1})^2}({x_2}{y_3} - {x_3}{y_2})\cdot $\\
				&&${({x_3}{y_4} - {x_4}{y_3})^2}$ \\
				(17) & $V(1,2,3)$ & $({x_1}{y_2} - {x_2}{y_1}){({x_2}{y_3} - {x_3}{y_2})^2}\cdot $\\
				&&${({x_3}{y_4} - {x_4}{y_3})^2}$ \\
				(18) & $V(1,2,3)$ & $({x_1}{y_2} - {x_2}{y_1}){({x_2}{y_3} - {x_3}{y_2})^2}\cdot $\\
				&&$({x_3}{y_4} - {x_4}{y_3}){({x_1}{y_4} - {x_4}{y_1})^3}$ \\
				(19) & $V(1,2,3)$ & $({x_1}{y_2} - {x_2}{y_1}){({x_2}{y_3} - {x_3}{y_2})^2}\cdot $\\
				&&${({x_3}{y_4} - {x_4}{y_3})^2}{({x_1}{y_4} - {x_4}{y_1})^2}$ \\
				(20) & $V(1,2,3)$ & $({x_1}{y_2} - {x_2}{y_1})({x_1}{y_3} - {x_3}{y_1})\cdot $\\
				&&${({x_1}{y_4} - {x_4}{y_1})^2}({x_2}{y_3} - {x_3}{y_2})$ \\
				(21) & $V(1,2,3)$ & $({x_1}{y_2} - {x_2}{y_1})({x_1}{y_3} - {x_3}{y_1})\cdot $\\
				&&${({x_1}{y_4} - {x_4}{y_1})^2}{({x_2}{y_3} - {x_3}{y_2})^3}$ \\
				(22) & $V(1,2,3)$ & $({x_1}{y_2} - {x_2}{y_1})({x_2}{y_3} - {x_3}{y_2})\cdot $\\
				&&$({x_3}{y_4} - {x_4}{y_3})({x_1}{y_4} - {x_4}{y_1})\cdot $\\
				&&$({x_1}{y_3} - {x_3}{y_1})$ \\
				(23) & $V(1,2,3)$ & ${({x_1}{y_2} - {x_2}{y_1})^2}{({x_2}{y_3} - {x_3}{y_2})^2}\cdot $\\
				&&$({x_3}{y_4} - {x_4}{y_3})({x_1}{y_4} - {x_4}{y_1})\cdot $\\
				&&$({x_1}{y_3} - {x_3}{y_1})$ \\
				(24) & $V(1,2,3)$ & $({x_1}{y_2} - {x_2}{y_1})({x_2}{y_3} - {x_3}{y_2})\cdot $\\
				&&${({x_3}{y_4} - {x_4}{y_3})^2}({x_1}{y_4} - {x_4}{y_1})\cdot $\\
				&&$({x_1}{y_3} - {x_3}{y_1})({x_2}{y_4} - {x_4}{y_2})$ \\
				\bottomrule
			\end{tabular}
		\end{table}		 		
		
		One expanded instances of SCAMI (\textit{SCAMI}(6)) in the form of rational expressions of shape-color moments is presented in equation (\ref{equ:expand}) as following.
		\begin{equation} \label{equ:expand}
		\frac{
			\left\{\begin{aligned}
			&2\cdot scU_{00002} \cdot scU_{02020}scU_{20200} - 4scU_{00002}scU_{02110}\cdot   \\
			&scU_{20110} + 2scU_{00002}scU_{02200}scU_{20020} - 4scU_{00002}\cdot   \\
			&scU_{11020}scU_{11200} + 4scU_{00002}scU_{11110}^2 - 4scU_{00011}\cdot   \\
			&scU_{02011}scU_{20200} + 4scU_{00011}scU_{02101}scU_{20110}+  \\
			&4scU_{00011}scU_{02110}scU_{20101} - 4scU_{00011}scU_{02200}\cdot   \\
			&scU_{20011} + 8scU_{00011}scU_{11011}scU_{11200} - 8scU_{00011}\cdot   \\
			&scU_{11101}scU_{11110} + 2scU_{00020}scU_{02002}scU_{20200}-  \\
			&4scU_{00020}scU_{02101}scU_{20101} + 2scU_{00020}scU_{02200}\cdot   \\
			&scU_{20002} - 4scU_{00020}scU_{11002}scU_{11200} + 4scU_{00020}\cdot   \\
			&scU_{11101}^2 + 4scU_{00101}scU_{02011}scU_{20110} - 4scU_{00101}\cdot   \\
			&scU_{02020}scU_{20101} - 4scU_{00101}scU_{02101}scU_{20020}+  \\
			&4scU_{00101}scU_{02110}scU_{20011} - 8scU_{00101}scU_{11011}\cdot   \\
			&scU_{11110} + 8scU_{00101}scU_{11020}scU_{11101} - 4scU_{00110}\cdot   \\
			&scU_{02002}scU_{20110} + 4scU_{00110}scU_{02011}scU_{20101}+  \\
			&4scU_{00110}scU_{02101}scU_{20011} - 4scU_{00110}scU_{02110}\cdot   \\
			&scU_{20002} + 8scU_{00110}scU_{11002}scU_{11110} - 8scU_{00110}\cdot   \\
			&scU_{11011}scU_{11101} + 2scU_{00200}scU_{02002}scU_{20020}-  \\
			&4scU_{00200}scU_{02011}scU_{20011} + 2scU_{00200}scU_{02020}\cdot   \\
			&scU_{20002} - 4scU_{00200}scU_{11002}scU_{11020} + 4scU_{00200}\cdot   \\
			&scU_{11011}^2
			\end{aligned}\right\}
		}
		{
			\left\{\begin{aligned}
			&scU_{00000}^2 \cdot (6 \cdot scU_{00002} \cdot scU_{00020} \cdot scU_{00200} -  \\
			&6 \cdot scU_{00002} \cdot scU_{00110}^2 - 6 \cdot scU_{00011}^2 \cdot scU_{00200} +  \\
			&12 \cdot scU_{00011} \cdot scU_{00101} \cdot scU_{00110} - 6 \cdot scU_{00020} \cdot \\
			&scU_{00101}^2)
			\end{aligned}\right\}
		}
		\end{equation}		
		
		The 24 simple and effective SCAMIs are constructed, which can be used as the basic descriptor in the process of analyzing and understanding images, which can be called \textit{SCAMI}24.
		
		\begin{equation}
			SCAMI24 = [SCAMI(1),...,SCAMI(24)]
		\end{equation}						
	\section{Experiments}\label{sec:experiments}
		In this section, the performance of SCAMI24 are evaluated against several approaches under both artificial and benchmark images. First, the invariant properties of these methods are theoretically discussed. Then, the invariance and discriminative are evaluated on both artificial and benchmark datasets. Finally, the results of experiments are presented.
		
		\subsection{Invariance Properties of Descriptors}				
		The SCAMI24 remains invariant under simultaneously shape and color affine transformations. The superiority of SCAMI24 for other methods are evaluated. Depending on the transformation modes established in Section \ref{sec:background}, there are difference invariance properties of descriptors corresponding to different degradations, such as invariance to geometric similarity, geometric affine, color linear, color diagonal, and color affine transformations. The descriptors used in the experiments are listed in following.
		
		\textit{Similar Moment Invariant(SMI).} Similar moment invariant are proposed by Hu in 1962 \cite{hu1962visual}. Hu used the geometric central moments to obtain seven geometric moment invariants with translational, rotation, and scaling invariance. SMI have invariance properties under similar transformation. They are calculated for the intensity images and no invariance for color deformations.
		
		\textit{Affine Moment Invariant(AMI).} Affine moment invariant (AMI) are proposed by Suk and Flusser \cite{suk2004graph} based on the algebraic invariant theory. AMI is invariant under geometric affine transformation, which is more robust than SMI when applying into real scene. They are also have no invariance for color deformations.
		
		\textit{Color Histogram(CH).} The color histogram is always used as the descriptor for the color images. It is the concatenation of three 1-D histograms based on the three channels of images. Color histogram ignores the shape structure in the images and has no invariance properties for color deformations.
		
		\textit{Transformed Color Histogram(TCH).} In order to achieve color invariance for color histogram, we can normalize the histogram distributions. Each channel is normalized independently by the mean and variance. The descriptor will be invariance to color transformation.
		
		\textit{Color Moments(CM).} 
		Stricker and Orengo proposed a method of color moments \cite{stricker1995similarity}, which is a simple and effective color descriptor. Since the color information is mainly distributed in the lower order moments,
		the first three order moments are sufficient to express the color distribution of the image. The color moments are invariant to color diagonal deformation.		
		
		\textit{Color Affine Moment invariants(CAMI).}
		Gong et al. proposed a kind of color affine moment invariants which are applicable to more complicated color variance \cite{gong2013moment}, but it has no invariance for shape transformations.		
		
		\textit{GPSO.}
		Mindru et al. proposed a kind of generalized color moment invariants\cite{mindru2004moment}, extending the moment invariants obtained by Lie group methods detailed in \cite{moons1995foundations,van1995vision}, which considers both shape and color deformations. The invariants are computed on different combinations of color channels. The invariance property is limited to color diagonal transformation.		
		
		\textit{AMI for Color Images(AMICI).}
		Suk and Flusser proposed another type of affine moment invariants for color images \cite{suk2009affine}, which is also depending on the generalized color moment invariants. Additional invariants can be computed on different combinations of color channels. Their invariance properties are limited to color diagonal transformation too.

		The invariance properties of descriptors are listed in the Table \ref{table:invpro}. 		
		\begin{table}[ht]
			\centering
			\small
			\caption{Invariance properties of descriptors}
			\label{table:invpro}
			\begin{tabular}{lccccc}
				\toprule
				& \begin{tabular}[c]{@{}c@{}}Shape \\ Similiar\end{tabular} & \begin{tabular}[c]{@{}c@{}}Shape\\ Affine\end{tabular} & \begin{tabular}[c]{@{}c@{}}Color \\ Linear\end{tabular} & \begin{tabular}[c]{@{}c@{}}Color\\  Diagonal\end{tabular} & \begin{tabular}[c]{@{}c@{}}Color \\ Affine\end{tabular} \\
				\midrule
				\textit{SMI}     & + & - & O & O & O \\
				\textit{AMI}     & + & + & O & O & O \\
				\textit{CH}      & O & O & - & - & - \\
				\textit{TCH} 	& O & O & + & + & - \\
				\textit{CM}      & O & O & - & - & - \\
				\textit{CAMI}    & - & - & + & + & + \\
				\textit{GPSO}    & + & + & + & + & - \\
				\textit{AMICI}   & + & + & + & + & - \\
				\textit{SCAMI}24   & + & + & + & + & + \\
				\bottomrule
			\end{tabular}
		\end{table}	
		
		Where invariance is represented by '+', lack of invariance is represented by '-'. 'O' represents the descriptor discards this kind information.
		
		From Table \ref{table:invpro} we can see that, the shape invariant descriptors, \textit{SMI} and \textit{AMI}, are computed on the gray-value images, which have shape invariant properties for difference geometric deformations, represented by '+'. But they discard the color information contained in the original image, represented by 'O'. For two identical objects with distinct colors which should be assigned to different categories, the shape descriptors do not have any discriminating power as to this case. Obviously, color information should be utilized by descriptor. 
		
		Color descriptors, such as \textit{CH}, \textit{TCH}, \textit{CM}, \textit{CAMI}, are make full use of the color information, which have different color invariant properties as shown in Tabel \ref{table:invpro}. Color descriptors based on histogram use the statical distribution of colors and shape consistency does not be considered. But some cases with the same color distribution may correspond to very different scenes. Other color descriptors also have the same problem. So shape consistency is also important for color descriptors. 
		
		The \textit{GPSO}, \textit{AMICI} and \textit{SCAMI}24 have both shape and color invariance properties. But just SCAMI24 are invariant for both shape and color affine transformation and can deal with more complicate situation, which will be demonstrated in experiments.		

		\subsection{Experiments of Synthetic Transformations}
		This experiment is designed to evaluate the invariance and robustness of the $SCAMI24$. The invariance under the shape affine transformation and color affine transformation is the inherent attribute of the $SCAMI24$.  
		
\begin{figure}
			\centering			
			\subfigure[]{ \label{fig:COREL1}
				\scalebox{0.11}{\includegraphics{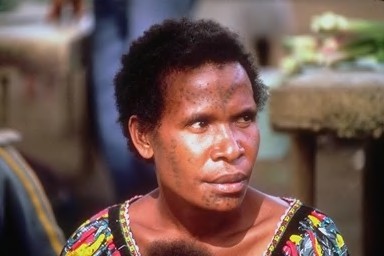}}}
			\subfigure[]{ \label{fig:COREL2}
				\scalebox{0.11}{\includegraphics{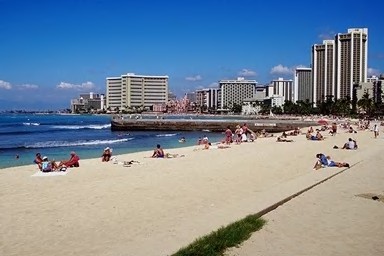}}}
			\subfigure[]{ \label{fig:COREL3}
				\scalebox{0.11}{\includegraphics{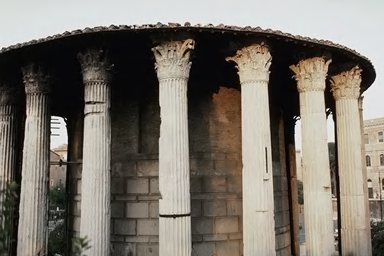}}}
			\subfigure[]{ \label{fig:COREL4}
				\scalebox{0.11}{\includegraphics{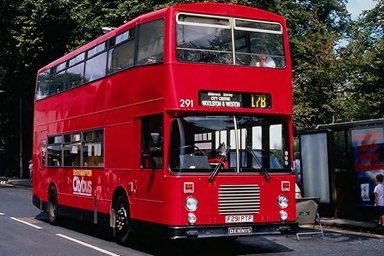}}}
			\subfigure[]{ \label{fig:COREL5}
				\scalebox{0.11}{\includegraphics{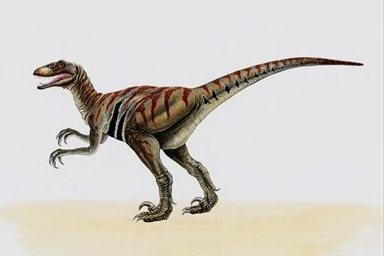}}}
			\subfigure[]{ \label{fig:COREL6}
				\scalebox{0.11}{\includegraphics{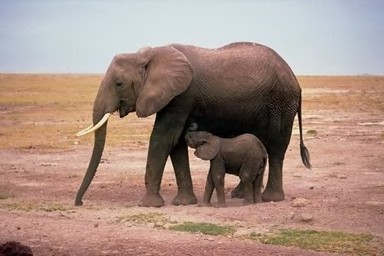}}}
			\subfigure[]{ \label{fig:COREL7}
				\scalebox{0.11}{\includegraphics{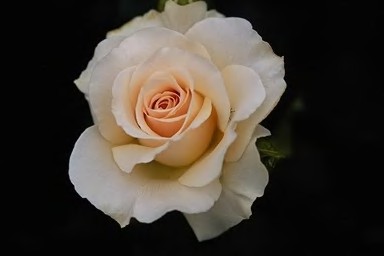}}}
			\subfigure[]{ \label{fig:COREL8}
				\scalebox{0.11}{\includegraphics{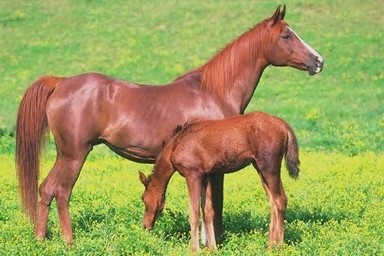}}}
			\subfigure[]{ \label{fig:COREL9}
				\scalebox{0.11}{\includegraphics{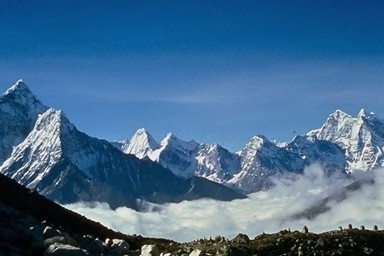}}}
			\subfigure[]{ \label{fig:COREL10}
				\scalebox{0.11}{\includegraphics{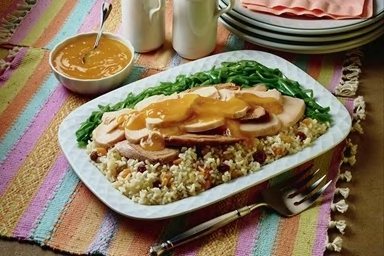}}}
			
			\caption{The original images from COREL1000.} \label{fig:COREL}
		\end{figure}

\begin{figure}
			\centering			
			\subfigure[]{ \label{fig:CORELss}
				\includegraphics[width=85mm]{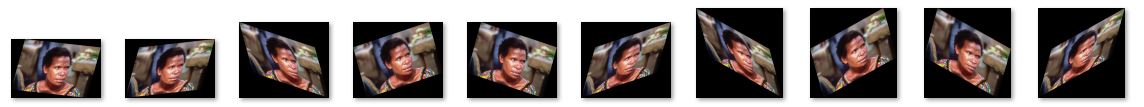}}
			\subfigure[]{ \label{fig:CORELsc}
				\includegraphics[width=85mm]{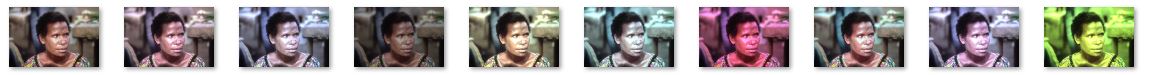}}
			\subfigure[]{ \label{fig:CORELssc}
				\includegraphics[width=85mm]{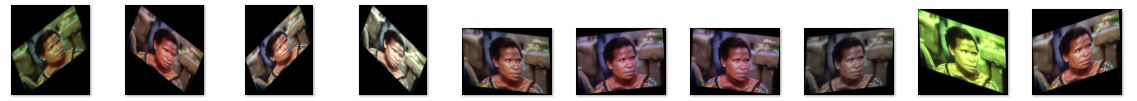}}
			
			\caption{Samples of transformations of original images: (a) samples of shape transformations; (b) samples of color transformations; (c) samples of shape-color dual-affine transformations.} \label{fig:CORELs}
		\end{figure}

		We select 10 images from each classes of COREL1000 dataset, as shown in Fig.(\ref{fig:COREL}), and make some geometric and color transformations artificially. A series of images are constructed, which are transformed from single image under the exact shape and color transformations, including translation, rotation, scaling, geometric affine, color diagonal and color affine transformations, as shown in Fig.(\ref{fig:CORELs}).
		Then three datasets are constructed. The dataset with single shape transformations is consist of 10 classes of 840 images. The dataset with single color transformations is consist of 10 classes of 240 images. And the dataset with shape-color dual-affine transformations is consist of 10 classes of 840 images.
		The transformations are all constructed with exact transformation matrix. 10\% images are selected randomly to be the training data and the rest make up the testing data.
		The Nearest Neighbor (NN) classifier based on the Chi-Square distance was performed as our classification method. The results are listed in Table \ref{table:exp2}.
		
		From the results, we can see that the \textit{SMI} achieves much lower accuracy than the \textit{AMI} on the dataset with single shape transformations. The datasets including affine transformations, however, the \textit{SMI} is just invariant to similar transformations. So it can be proved that the transformation model is important for the invariant designing. 
		
		The color descriptor \textit{CH} and \textit{TCH} achieve perfect performance on the dataset with single shape transformations,
		because they just ignore all the space structure of pixels. So it can perform well for the images with exact shape transformation. But some cases with the same color distribution may correspond to very different shapes.
		For the datasets with color transformation. The two descriptors perform not so well.
		
		For \textit{CAMI}, it performs pretty better than most other descriptors on the dataset with single color transformation. Because it has invariant property for color affine transformation. But for the datasets with shape transformations, it performs not well. So \textit{CAMI} is just an color invariant descriptor and is sensitive to shape transformations.
		
		\begin{table}
			\centering
			\caption{Classification accuracy of different descriptors on artificial datasets based on COREL1000 }
			\small
			\label{table:exp2}
			\begin{tabular}{llll}
				\toprule
				& \multicolumn{1}{c}{shape} & \multicolumn{1}{c}{color} & \multicolumn{1}{c}{shape-color} \\
				\midrule
				\textit{SMI}   & 55.0265                   & 51.3889                   & 41.7376                             \\
				\textit{AMI}   & \textbf{100}                       & 47.4444                   & 41.5344                             \\
				\textit{CH}    & \textbf{100}                       & 46.7593                   & 67.5926                             \\
				\textit{TCH}   & \textbf{100}                       & 43.5158                   & 75                                  \\
				\textit{CM}    & 72.0899                   & 71.2963                   & 53.1746                             \\
				\textit{CAMI}  & 12.3016                   & 79.6296                   & 14.6825                             \\
				\textit{GPSO}  & 88.0952                   & 69.9074                   & 62.6984                             \\
				\textit{AMICI} & 91.0053                   & 74.5370                   & 82.4074                             \\
				\textit{SCAMI24} & \textbf{100}                       & \textbf{89.8148}                   & \textbf{91.4021}                             \\
				\bottomrule				
			\end{tabular}
		\end{table}
		
		For descriptors invariant to both shape and color, \textit{GPSO}, \textit{AMICI} and \textit{SCAMI}24, they are all perform well on all the three datasets. \textit{GPSO} and　\textit{AMICI} perform not so well on datasets with color transformations, because they have no invariant to the color affine transformations.
		\textit{SCAMI}24 achieves the best performance, because it is invariant to color affine transformations. And it also remains invariance to the shape affine transformation. The superiority of \textit{SCAMI}24 is obvious.		
		
\begin{figure}
		       	\centering
		       	
		       	\subfigure[]{ \label{fig:pr1}
			       	\includegraphics[width=85mm]{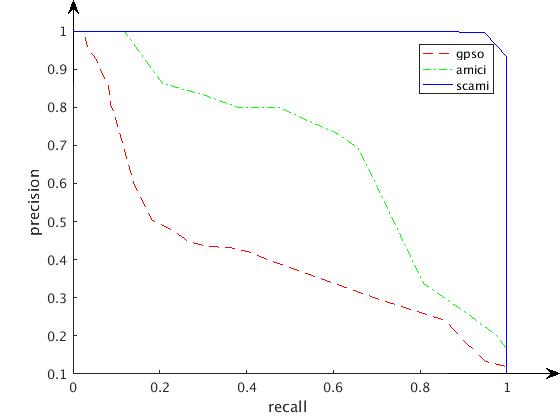}}
		       	\subfigure[]{ \label{fig:pr2}
		       		\includegraphics[width=85mm]{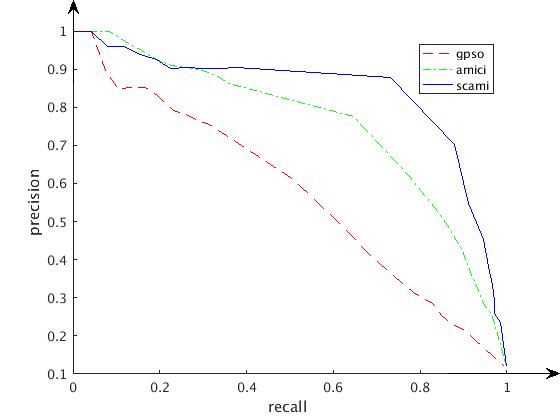}}
		       	\subfigure[]{ \label{fig:pr3}
		       		\includegraphics[width=85mm]{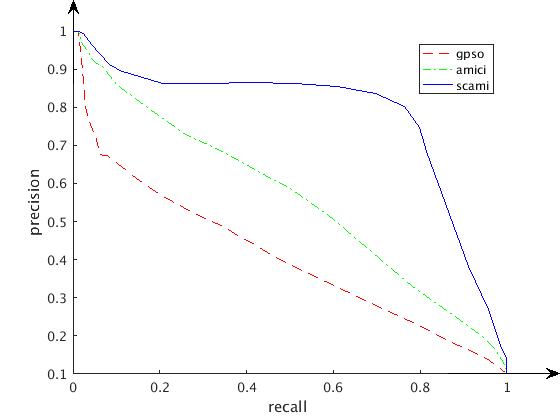}}
		       	
		       	\caption{PR Curves: (a) PR Curves of Dataset with Shape Affine Transformations; (b) PR Curves of Dataset with Color Affine Transformations; (c) PR Curves of Dataset with Shape-color Dual-affine Transformations.} \label{fig:pr}
		\end{figure}

		For a comprehensive research for the \textit{SCAMI}24, we also conduct image retrieval experiments on the three datasets.
		The Precision-Recall curve is chosen to evaluate the results. The result are shown on Fig. (\ref{fig:pr}). 
		
		\begin{equation}
			\begin{array}{l}
			Precision = \frac{{|\{ relevant{\rm{ }}images\}  \cap \{ retrieved{\rm{ }}images\} |}}{{|\{ retrieved{\rm{ }}images\} |}}\\
			Recall = \frac{{|\{ relevant{\rm{ }}images\}  \cap \{ retrieved{\rm{ }}images\} |}}{{|\{ relevant{\rm{ }}images\} |}}
			\end{array}
		\end{equation}		
		
		It is shown that for this  dataset, our $SCAMI24$ performs better than the other two descriptors.
		
		\subsection{Experiments of Real Scene Transformations on ALOI Dataset}\label{subsec:exp3}
		Image classification experiments are conducted on the Amsterdam Library of Object Images Database (ALOI)\cite{geusebroek2005amsterdam}, which contains images of 1000 objects taking from various imaging conditions. The images in the dataset are all the black background except the objects. The imaging condition  difference includes the illumination colors, illumination arrangements and viewpoints. The illumination and viewpoint changes can be approximated by affine transformation. The ALOI is very suitable for the evaluation of the global descriptors.

\begin{figure}
			\centering			
			\subfigure[]{ \label{fig:ALOI1}
				\includegraphics[width=85mm]{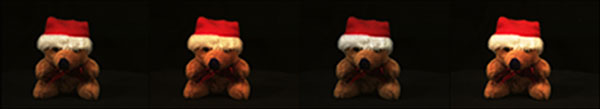}}
			\subfigure[]{ \label{fig:ALOI2}
				\includegraphics[width=85mm]{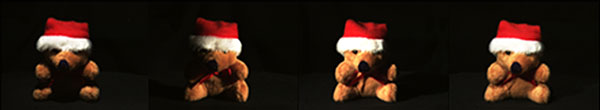}}
			\subfigure[]{ \label{fig:ALOI3}
				\includegraphics[width=85mm]{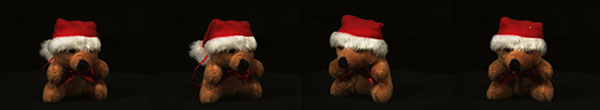}}		
			\caption{Sample images from ALOI database: (a) samples of image under different illumination colors; (b) samples of image under different illumination arrangements; (c) samples of image with different viewpoints.} \label{fig:ALOI}
		\end{figure}

		The examples of ALOI dataset are shown in Fig. (\ref{fig:ALOI}). We choose 10 classes as our dataset. Each class, there are 12 images with light color changes and 11 images with viewpoint changes. To extend the dataset, we apply artificial transformations into the images. Then we get four datasets CC, SS, CS and SC. CC contains images with light color changes and artificial color changes. SS contains images with viewpoint changes and artificial shape changes. CS contains images with light color changes and artificial shape changes. And SC contains images with viewpoint changes and artificial color changes.		
		For the CC, it contains 1200 images with only color deformation. SS contains 840 images with only shape deformation. CS and SC contain images with both shape and color deformations, the number of images are 840 and 1100. 
		
		The Nearest Neighbor (NN) classifier based on the Chi-Square distance was performed as our classification method. Four experiments are conducted separately. The results are listed in Table \ref{table:exp3}.
		\begin{table}[ht]
			\centering
			\caption{Classification accuracy of different descriptors on datasets based on ALOI }
			\label{table:exp3}
			\begin{tabular}{llll}
				\toprule
				& \multicolumn{1}{c}{\textit{GPSO}} & \multicolumn{1}{c}{\textit{AMICI}} & \multicolumn{1}{c}{\textit{SCAMI}24} \\
				\midrule
				CC & 94.1667                  & 86.4815                   & \textbf{95.6481}                   \\
				SS & 83.3333                  & 91.0053                   & \textbf{96.2963}                   \\
				CS & 76.7196                  & 94.3122                   & \textbf{96.9577}                   \\
				SC & 85.8886                  & 89.0909                   & \textbf{94.1414}                   \\
				\bottomrule
			\end{tabular}
		\end{table}	
		It can be seen that the $SCAMI24$ performs best among the three descriptors. 
		
		From above experimental results, it has been proved that, as a kind of global descriptor, $SCAMI24$ performs better than the other global descriptors and robust for the real scenery images classification.
		\subsection{Results} 		
		For images with artificial transformations, $SCAMI24$ performs well on  both image classification and image retrieval experiments. And for images got from the real scenery, $SCAMI24$ also performs well in the image classification experiments. The $SCAMI24$ is recommended as a global descriptor in image processing and understanding.
		
	\section{Conclusion}
		A kind of naturally combined shape-color affine moment invariants, called SCAMIs are proposed in this paper, which unify shape and color factors together and are invariant to affine transformations of both shape and color simultaneously. There are several benefits for the SCAMIs. Primarily, it is the first time to directly derive invariants to dual affine transformations of shape and color, which are the most complicated linear transformations. Secondly, there is no requirement for the selection of weights to combine two features. Most of all, the method used here is a complete framework which can be extended into high order and higher dimension conveniently. In additional, a set of SCAMIs, which called $SCAMI24$, is constructed and recommended as image descriptors or features.

		By definition, SCAMIs are global invariant. The extension of them to some kinds of local descriptors or to more general transformation than affine is just under way.

	\section*{References}
	\bibliographystyle{elsarticle-num}
	\bibliography{scami_bib}
	
\end{document}